\documentclass{article}
\usepackage{cite} 

\usepackage{arxiv}

\usepackage[utf8]{inputenc} 
\usepackage[T1]{fontenc}    
\usepackage{hyperref}       
\usepackage{url}            
\usepackage{booktabs}       
\usepackage{amsfonts}       
\usepackage{nicefrac}       
\usepackage{microtype}      
\usepackage{cleveref}       
\usepackage{lipsum}         
\usepackage{graphicx}
\usepackage{natbib}
\usepackage{doi}

\title{An accuracy improving method for advertising click through rate prediction based on enhanced xDeepFM model}

\newif\ifuniqueAffiliation
\uniqueAffiliationtrue

\ifuniqueAffiliation 
\author{%
    Xiaowei Xi\\
    School of Dig Data\\
    Zhuhai College of Science and Technology\\
    Zhuhai, China, 519041\\
    \texttt{xxw030224@stu.zcst.edu.cn} \\
    \And
    Song Leng\\
    School of Computer Science\\
    Zhuhai College of Science and Technology\\
    Zhuhai, China, 519041\\
    \texttt{lengsong2022@zcst.edu.cn} \\
    \And
    Yuqing Gong\\
    School of Computer Science\\
    Zhuhai College of Science and Technology\\
    Zhuhai, China, 519041\\
    \texttt{gongyuqing1975@zcst.edu.cn} \\
    \And
    Dalin Li\\
    School of Computer Science\\
    Zhuhai College of Science and Technology\\
    Zhuhai, China, 519041\\
    \texttt{lidalin@zcst.edu.cn} \\
}
\else

\newbox{\orcid}\sbox{\orcid}{\includegraphics[scale=0.06]{orcid.pdf}} 

\author[1]{%
    \href{https://orcid.org/0000-0000-0000-0000}{\usebox{\orcid}\hspace{1mm}David S.~Hippocampus\thanks{\texttt{hippo@cs.cranberry-lemon.edu}}}%
}
\author[1,2]{%
    \href{https://orcid.org/0000-0000-0000-0000}{\usebox{\orcid}\hspace{1mm}Elias D.~Striatum\thanks{\texttt{stariate@ee.mount-sheikh.edu}}}%
}
\affil[1]{Department of Computer Science, Cranberry-Lemon University, Pittsburgh, PA 15213}
\affil[2]{Department of Electrical Engineering, Mount-Sheikh University, Santa Narimana, Levand}
\fi

\renewcommand{\shorttitle}{\textit{arXiv} Template}
\hypersetup{
pdftitle={A template for the arxiv style},
pdfsubject={q-bio.NC, q-bio.QM},
pdfauthor={David S.~Hippocampus, Elias D.~Striatum},
pdfkeywords={First keyword, Second keyword, More},
}

\begin{document}
\maketitle

\begin{abstract}
Advertising click-through rate (CTR) prediction aims to forecast the probability that a user will click on an advertisement in a given context, thus providing enterprises with decision support for product ranking and ad placement. However, CTR prediction faces challenges such as data sparsity and class imbalance, which adversely affect model training effectiveness. Moreover, most current CTR prediction models fail to fully explore the associations among user history, interests, and target advertisements from multiple perspectives, neglecting important information at different levels. To address these issues, this paper proposes an improved CTR prediction model based on the xDeepFM architecture. By integrating a multi-head attention mechanism, the model can simultaneously focus on different aspects of feature interactions, enhancing its ability to learn intricate patterns without significantly increasing computational complexity. Furthermore, replacing the linear model with a Factorization Machine (FM) model improves the handling of high-dimensional sparse data by flexibly capturing both first-order and second-order feature interactions. Experimental results on the Criteo dataset demonstrate that the proposed model outperforms other state-of-the-art methods, showing significant improvements in both AUC and Logloss metrics. This enhancement facilitates better mining of implicit relationships between features and improves the accuracy of advertising CTR prediction.

\end{abstract}

\keywords{Advertising click-through rate; Deep learning; XDeepFM model; Multi-head attention mechanism; Factorization machine}

\section{Introduction}
In today's data-driven and information-rich era, the Internet has become a crucial channel for people to acquire information, communicate, and conduct business activities. With the continuous expansion of online advertising scale and the accumulation of user behavior data, driven by economic development, both academia and industry are increasingly focusing on the field of advertising click-through rate (CTR)\cite{sinclair2020cracking}. CTR prediction \cite{sundaram2020power} has emerged as an important research topic in digital marketing. It involves predicting the likelihood that a user will click on an advertisement after viewing it by analyzing relevant ad features and user behavior data, thereby improving ad click performance \cite{gudipudi2023improving}. This prediction process is not only significant for advertisers' decision-making but also optimizes advertising placement strategies \cite{gengler1995consumer}, enhancing the utilization of advertising resources. CTR prediction is commonly used in the online advertising \cite{evans2009online} industry and serves as one of the core bases for ad placement, holding great practical significance. It is also a key indicator used by advertisers to measure advertising effectiveness \cite{wiles1991review}. Improving the accuracy of CTR prediction is crucial for maximizing the benefits of advertisers, platform providers, and users \cite{wuisan2023maximizing}. An increase in CTR accuracy can lead to higher revenue for advertisers and platforms and, from the user's perspective, help enhance user experience and increase user engagement \cite{rong2019platform}.

Traditional CTR prediction algorithms have played an important role in the advertising field. Their simplicity and effectiveness offer certain advantages in solving ad placement problems, including but not limited to: Logistic Regression (LR) models advertising features using linear weights, demonstrating flexibility in handling various types of features \cite{hosmer2013applied}. Gradient Boosting Decision Tree (GBDT) automates feature engineering by exploring entropy-based relationships among features, significantly reducing the complexity of manual feature selection \cite{feng2018multi}. Factorization Machine (FM), by representing features as combinations of factors, effectively captures interactions between features, making it particularly suitable for scenarios with high data sparsity and showcasing exceptional performance \cite{rendle2010factorization}. The GBDT+LR model combines the feature extraction capability of GBDT with the predictive power of LR, addressing the limitations of standalone models in managing both nonlinear feature interactions and inference efficiency \cite{he2014practical}. Specifically, GBDT generates high-order feature combinations, which are then used as input for LR to make predictions. Multilayer Logistic Regression (MLR) extends the traditional LR with a multilayer structure, thereby enhancing its ability to capture complex nonlinear relationships among features \cite{gai2017learning}.However, traditional CTR prediction algorithms have weak generalization ability under new data distributions and scenarios, which may lead to performance degradation. For highly correlated features, these algorithms are prone to overfitting, instability, and reduced model interpretability. When dealing with large-scale data, the efficiency and performance of traditional algorithms are limited, and they struggle to effectively capture the associations between ads and user contexts, thereby limiting the model's comprehensiveness.

In recent years, deep learning has achieved remarkable success in fields such as computer vision and natural language processing\cite{wiriyathammabhum2016computer}. Its application in click-through rate (CTR) prediction for advertisements has also gained significant traction, driving improvements in both prediction accuracy and efficiency. Compared to traditional machine learning methods\cite{mahesh2020machine}, deep learning models exhibit superior feature representation capabilities, enabling the automatic extraction and modeling of complex nonlinear relationships within data. This advantage is particularly critical for capturing user behavior patterns, which makes deep learning-based CTR prediction algorithms stand out with significantly enhanced performance, far surpassing traditional methods.In practical applications, Long Short-Term Memory (LSTM) \cite{hochreiter1997long} networks have been widely utilized for modeling the temporal dependencies of user click behaviors, leveraging their superior sequence data modeling capabilities to improve prediction performance. Convolutional Neural Networks (CNNs)\cite{o2015introduction}, through convolutional operations, extract local patterns and efficiently handle high-dimensional sparse features in advertisements, such as text and images. Feedforward Neural Networks (FNNs)\cite{bebis1994feed}, which combine the feature interaction capabilities of Factorization Machines (FM) with the nonlinear expressive power of deep networks, achieve efficient transformations from sparse data to dense representations via multi-layer fully connected networks.Building on these foundations, Google's research team proposed the Wide \& Deep model\cite{cheng2016wide}, which integrates shallow logistic regression and deep neural networks (DNNs) to balance memorization of specific features and generalization capability. Subsequently, Guo et al.\cite{guo2017deepfm} optimized this model by replacing logistic regression with Factorization Machines (FM) and sharing the output of the embedding layer, thereby introducing the DeepFM model. This improvement reduced the complexity of feature engineering and enhanced feature interaction capabilities. However, DeepFM still exhibited limitations in modeling high-order feature interactions.To address these shortcomings, Lian et al.\cite{lian2018xdeepfm}proposed the xDeepFM model, combining the strengths of DeepFM and Deep \& Cross Networks (DCN)\cite{wang2017deep}. By introducing the Compressed Interaction Network (CIN), xDeepFM explicitly models arbitrary high-order feature interactions, significantly enhancing its ability to capture complex user behavior patterns. These advancements illustrate that the application of deep learning in CTR prediction not only improves model performance but also accelerates the evolution of advertising systems toward intelligent solutions.

xDeepFM fully leverages the advantages of deep learning. The model adopts a deep and cross network structure to jointly model user behavior and user interest evolution, enabling the model to learn both explicit and implicit high-order feature interactions. In this structure, the cross network is responsible for capturing the interaction relationships between user information, while the deep network is used to mine deep features, thereby greatly improving the accuracy of CTR prediction. However, the linear part of the original xDeepFM model mainly relies on shallow linear feature representations, limiting the model's ability to model nonlinear relationships \cite{wang2019survey} and high-order feature interactions. It overlooks the varying degrees of relevance that different low-order feature interactions have with the prediction target and cannot accurately capture differences and trends in user behavior across different time periods, which hinders high-order feature learning.

To address these issues, this paper integrates a multi-head attention mechanism into the CTR prediction model based on the xDeepFM model, allowing the model to enhance interactive learning between features and better capture the correlations and weights among different features. Additionally, the FM model is used to replace the linear model component, extracting first-order features and second-order features formed by pairwise combinations of first-order features, learning interaction factors between features, thereby enabling the model to better handle high-dimensional discrete features and improving its ability to model highly sparse data. The FM model offers considerable flexibility in parameter selection and adjustment for fitting low-order features. Experiments demonstrate that introducing the multi-head attention mechanism and using the FM model to replace the linear model component can enhance the model's ability to model complex feature relationships and perform better on high-dimensional discrete data, thus improving overall prediction performance.

The contributions of this paper are as follows:
(1)	Introduced a multi-head attention mechanism into the xDeepFM model to improve model performance.
(2)	Replaced the original LR model in the xDeepFM model with the FM model.
(3)	Achieved significant improvements in prediction accuracy compared to other similar models.

The structure of this paper is arranged as follows: Section 2 introduces the proposed improved CTR prediction model and provides a detailed description of each component; Section 3 presents the experimental results; Section 4 discusses the findings and concludes the paper.

\section{CTR Prediction Model Based on Improved xDeepFM Model}
Although the deep learning recommendation model xDeepFM, used for CTR prediction as a powerful feature interaction framework, has achieved remarkable results, several issues remain. First, the xDeepFM model uses the LR model to extract first-order features, ignoring the varying importance of different features to the target vector. Second, although xDeepFM considers the FM component when processing feature interactions, further optimization is needed in handling feature sparsity to better capture associations between rare features. Additionally, the modeling of temporal correlations is insufficient, particularly in time-sensitive tasks such as CTR prediction, the model may not accurately capture differences and trends in user behavior across different time periods. Therefore, this paper proposes a CTR prediction model based on an improved xDeepFM model, integrating a multi-head attention mechanism into the xDeepFM model and replacing the linear model component with the FM model to address the aforementioned issues.

\label{sec:headings}

\subsection{Overall Structure of the Model}
This paper presents an advertising click-through rate (CTR) prediction method based on an enhanced xDeepFM model, as illustrated in Figure 1. The proposed model incorporates critical improvements to the original xDeepFM framework. Firstly, it introduces a multi-head attention mechanism to more effectively capture the correlations between users' historical behavior sequences and the advertisements to be predicted. This enhancement allows for the extraction of users' latent interests while mitigating the interference of irrelevant historical behaviors on the prediction outcomes. Secondly, the model replaces the original linear component with a Factorization Machine (FM) to better learn low-order feature interactions, thereby enhancing the model's expressive capability. Specifically, the raw features are one-hot encoded \cite{okada2019efficient} and then passed through an embedding layer to compress and map the high-dimensional sparse features into low-dimensional dense vectors. Subsequently, by integrating FM, Compressed Interaction Network (CIN), and a deep neural network module augmented with a multi-head attention mechanism, the model is capable of explicitly learning high-order feature interactions. Additionally, through multi-level nonlinear transformations, it intricately and implicitly extracts the underlying relationships among users' histories, interests, and target advertisements \cite{zhang2021deep}. Finally, the outputs from each module are normalized using a sigmoid function, and parameters are adjusted through appropriate optimization methods to generate the final CTR prediction results.

\begin{figure}[h!]
    \centering
    \includegraphics[width=0.8\textwidth]{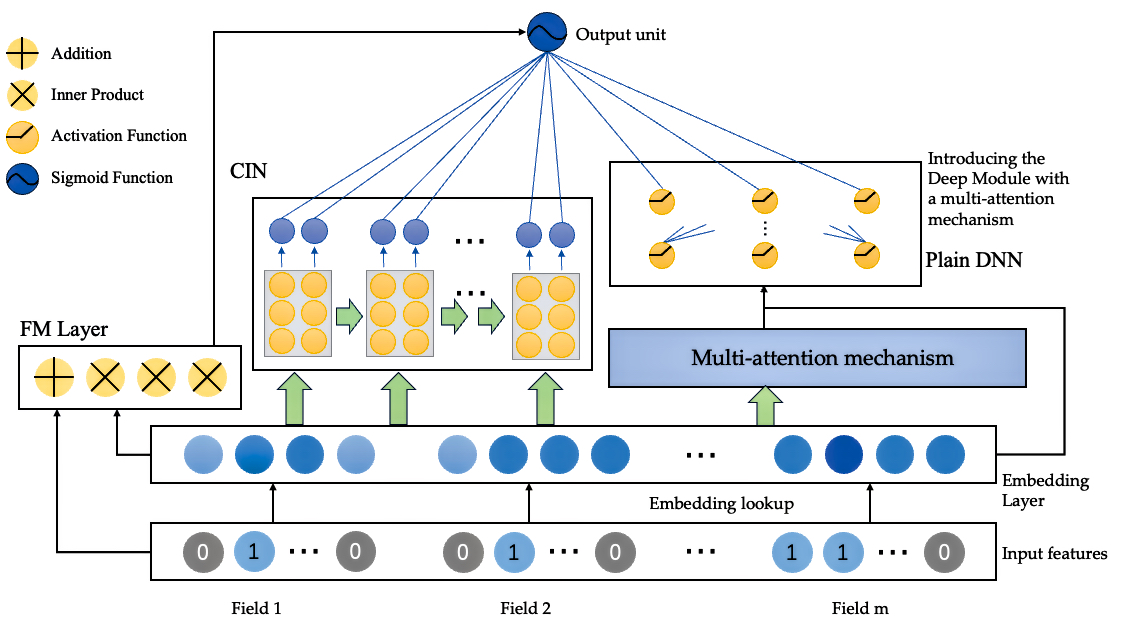}
    \caption{Overall architecture of the improved xDeepFM model incorporating the multi-head attention mechanism and FM}
    \label{fig:example}
\end{figure}

\subsection{Factorization Machine Model}
To overcome the LR model's limitation in capturing feature interrelationships, we introduced the Factorization Machine (FM) model, which effectively learns complex feature interactions. By incorporating factorized parameters, FM maps each feature into a low-dimensional latent vector space, enabling the model to automatically learn and model complex feature interactions. This approach not only enhances the model's efficiency in handling feature interactions but also significantly improves overall predictive performance, overcoming many of the inherent deficiencies of traditional LR models. The formula is as follows:

\begin{equation}
\hat{y}(\mathbf{x}):=w_0+\sum_{i=1}^n w_i x_i+\sum_{i=1}^n \sum_{j=i+1}^n\left\langle\mathbf{v}_i, \mathbf{v}_j\right\rangle x_i x_j
\end{equation}

Style: \(\hat{y}(\mathbf{x})\) denotes the predicted output of the model.\(w_{0}\) indicates a bias term; \(w_{i}\) denotes the weight of the \(\mathbf{i}^{th}\) feature; \(x_{i}\) denotes the value of the \(\mathbf{i}^{th}\) feature; \(x_{j}\) denotes the value of the \(\mathbf{j}^{th}\) feature; The dot product of vectors \(\langle \mathbf{v}_i, \mathbf{v}_j \rangle\) denotes the similarity of two features in the hidden space.

\subsection{Compressed Interaction Network}
Figure 2 illustrates the structural design of the Compressed Interaction Network (CIN). The CIN model draws inspiration from convolutional neural networks in the field of computer vision, specifically adopting explicit methods for interactive feature learning, with a particular emphasis on the fine modeling of high-order feature interactions. Its core mechanism employs a series of sub-networks that utilize convolution operations to effectively capture and process feature crossings of varying levels and magnitudes, thereby significantly enhancing the model's ability to understand and express complex feature relationships.

There are \(N\) features, each mapped to a \(D\)-dimensional embedding vector. These \(N\) embedding vectors can be represented in matrix form as \(\mathbf{X}^{0} \in \mathbb{R}^{N \times D}\). The \(\mathbf{k}^{th}\) hidden layer of the network is denoted as \(\mathbf{X}^{k}\), and the neuron output of the current layer is computed based on the output of the previous layer and the embedding vector, which is computed as:

\begin{equation}
X_{h, *}^k=\sum_{i=1}^{H_{k-1}} \sum_{j=1}^N W_{i j}^{k, h}\left(X_{i, *}^{k-1} \circ X_{j, *}^0\right)
\end{equation}

Style: \(X_{h, *}^{k}\) denotes the output vector of the \(h\)th row of neurons in the \(k\)th layer; \(X_{i, *}^{k-1}\) denotes the output vector of the neuron in row \(i\) in layer \(k-1\); \(X_{j, *}^{0}\) denotes the output vector of the \(j\)th row of neurons in the input layer; \(W_{ij}^{k, h}\) denotes the parameter in row \(i\) and column \(j\) of the parameter matrix; \(\circ\) represents the multiplication of the corresponding dimension elements between the vectors. Next, the sub-networks are summed up by the sum-pooling operation to generate a summarized feature representation, which is computed as:

\begin{equation}
p_i^k=\sum_{j=1}^D X_{i, j}^k
\end{equation}

Get a pooling vector \(p^{k} = \left[ p_{1}^{k}, p_{2}^{k}, \ldots, p_{H_{k}}^{k} \right]\). The pooling vectors of the \(k\) layers are then concatenated to get \(p^{+} = \left[ p^{1}, p^{2}, \ldots, p^{k} \right]\).

\begin{figure}[h!]
    \centering
    \includegraphics[width=0.8\textwidth]{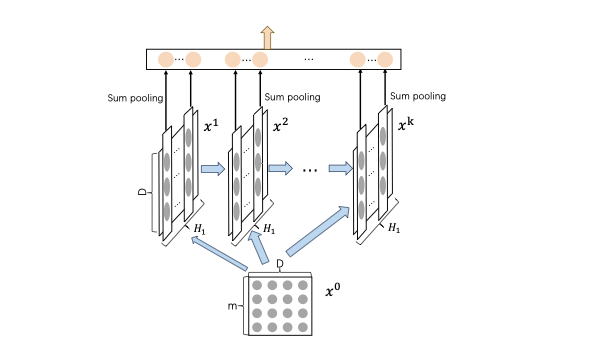}
    \caption{The Architecture of CIN}
    \label{fig:example}
\end{figure}

\subsubsection{Multi-head Attention Layer}
Inspired by the research of Song et al. \cite{song2019autoint}, we combine the attention mechanism with the xDeepFM model, introducing a preposed multi-head attention layer into its architecture, as shown in Figure 3. This attention mechanism adopts the multi-head attention from the Transformer model \cite{vaswani2017attention} and incorporates a residual structure \cite{wang2022spatial} to ensure that the effective information of the original embedding vectors is retained during feature interactions. By introducing a preposed multi-head attention mechanism, the model can automatically identify relevant features, capture complex interaction patterns, and map single features into multiple different subspaces, thereby constructing high-order features with greater expressive ability. This mechanism optimizes model training effectiveness without significantly increasing training complexity.
As a preposed module, the multi-head attention layer can finely analyze the relationships between original features and adaptively capture complex feature interaction patterns. Combined with the residual structure, the input embedding vectors are added to the attention outputs to form residual connections, ensuring that the original information of the input features is retained while capturing complex interactions. This retention mechanism ensures that the model does not lose basic information when learning new features. The outputs after residual connections undergo layer normalization to stabilize the training process and improve the model's robustness, providing rich and optimized feature inputs for the DNN layer. On this basis, the DNN further extracts and combines features using its powerful expressive ability, generating higher-level feature representations. The calculation mechanism of multi-head attention is as follows:
In the multi-head attention mechanism, the input vectors are first linearly transformed to form the Query matrix (Q), Key matrix (K), and Value matrix (V). Then, the dot product of the Query matrix (Q) and the Key matrix (K) is calculated to obtain attention scores. To prevent gradient issues in the subsequent Softmax calculation, the dot product result is appropriately scaled, usually by dividing by the square root of the dimensionality of the key vectors.

\begin{equation}
{ Attention }(Q, K, V)={softmax}\left(\frac{Q K^T}{\sqrt{d_k}}\right) V 
\end{equation}

The scaled dot product computation was transformed into an attention weight distribution using the Softmax function. Subsequently, the attention weights were multiplied with the value matrix (V) to obtain the output of the multiple attention. The computational equation is given below:

\begin{equation}
{ MultiHead }(Q, K, V)={ Concat }\left({ head }_1, \cdots, { head }_h\right) W^o
\end{equation}

\begin{equation}
{head}_i={ Attention }\left(Q W_i^Q, K W_i^K, V W_i^V\right)
\end{equation}

\begin{figure}[h!]
    \centering
    \includegraphics[width=0.8\textwidth]{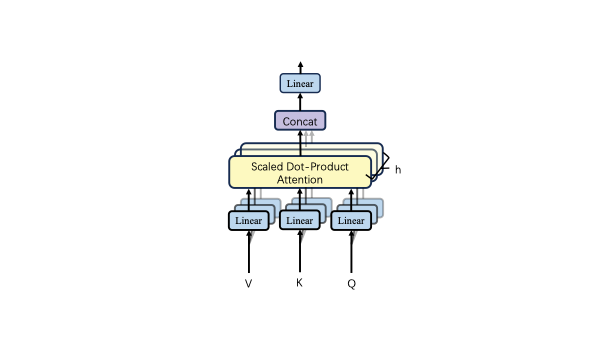}
    \caption{Structural diagram of multi-head attention mechanism}
    \label{fig:example}
\end{figure}

\subsubsection{Fusion Output Layer}
This module fuses the high-order feature interaction components processed by the FM, CIN, and Deep modules integrated with the multi-head attention mechanism. The FM model is used to learn second-order relationships between features; the CIN model further extends feature interactions, focusing on learning higher-order interaction relationships, effectively modeling large-scale high-order feature interactions without adding excessive parameters, thereby enhancing the model's nonlinear expressive ability. The multi-head attention mechanism allows the model to focus on the importance of features from different attention perspectives, helping it better capture the correlations and importance among different features. By integrating the Deep module with the multi-head attention mechanism, the model can more comprehensively consider feature interactions. Finally, the outputs processed by these modules are combined to produce the final CTR prediction result, thereby improving the model's expressive ability.

In this, a Sigmoid function is used to control the output of the model. On the one hand, the model includes both low-order and high-order feature interactions; on the other hand, the model includes both implicit and explicit feature interactions. The resulting output unit is computed as:

\begin{equation}
\hat{y}=\sigma\left(w_{F M}^T y_{F M}+w_{C I N}^T p^{+}+w_{d n n}^T x_{d n n}^k+b\right)
\end{equation}

Style: \(\sigma\) denotes the Sigmoid function. \(y_{FM}\), \(p^{+}\), and \(x_{dnn}^{k}\) denote the outputs of the FM model, the CIN model, and the Deep module, respectively. \(w\) and \(b\) denote the learnable parameters. 

This fusion improves the predictive power and accuracy of the model by integrating the strengths of different models and synthesizing the information learned from these modules, which can capture the complex interactions between features more efficiently and help to predict the user's clicking behavior more accurately.

\section{Experiments}
\subsection{Experimental Dataset}
The dataset used in this paper is the Criteo dataset, a public dataset released by Criteo in 2014 for CTR prediction. The records are sorted by time, containing 45 million samples over 7 days, with 13 continuous features and 26 categorical features. The distribution of data feature types in the dataset is shown in Figure 4.
This dataset is large-scale, with a high degree of sparsity, embodying the two main characteristics of massive and sparse data in current internet advertising CTR data. Stratified sampling was used to select samples, and 500,000 samples were used for training to avoid inconsistencies in the overall positive and negative sample ratio with the original data, ensuring that the experimental results are not affected.

\begin{figure}[h!]
    \centering
    \includegraphics[width=0.8\textwidth]{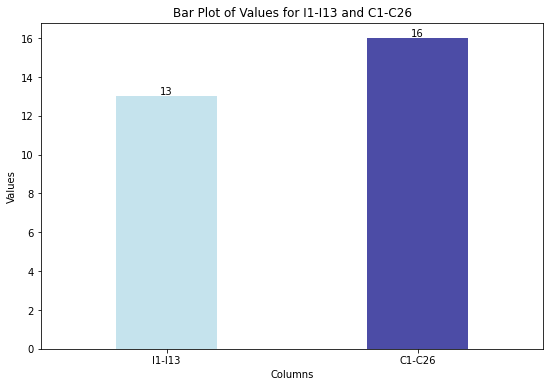}
    \caption{Distribution of data feature types}
    \label{fig:example}
\end{figure}

\subsection{Evaluation Metrics}
CTR prediction is essentially a binary classification problem \cite{deng2021deeplight}. To evaluate the model's effectiveness, this paper uses AUC and Logloss as evaluation metrics.
AUC measures the ability of the model to rank the prediction results \cite{lobo2008auc}. Specifically, it is the area under the ROC curve \cite{bradley1997use}. In binary classification, AUC measures the re-lationship between the model's True Positive Rate and False Positive Rate under different thresholds, takes into account the model's performance under different classification thresholds, and is more robust. The specific calculation formula is as follows:

\begin{equation}
A U C=\frac{\sum_{ {ins } \in { positiveclass }} {rank}_{{ins }}-\frac{M \times(M+1)}{2}}{M \times N}
\end{equation}

Style: rank denotes the ranking of the sample. \(\sum_{i \in {positive class}} {rank}_{i}\) represents the sum of the rankings in the positive category sample; \(M\) represents the number of positive category samples and \(N\) represents the number of negative category samples.

Logloss measures the difference between the model's predicted probabilities and the true labels \cite{bishop1995neural}. Specifically, it measures how well the model's predicted probabilities for each category match the actual category labels. It is a common metric used to evaluate the performance of classification models, especially in binary and multicategory classification problems. Logloss is widely used as a metric to evaluate model performance in binary classification tasks. The specific formula is as follows:

\begin{equation}
{ Logloss }=-\frac{1}{N} \sum_{i=1}^n y \log (p)+(1-y) \log (1-p)
\end{equation}

Style: \(N\) indicates the total number of samples in the data set; \(y\) denotes the actual category label, which can only take two values, 0 or 1. Where 0 means not clicked and 1 means clicked; \(p\) denotes the probability that the sample predicted by the model belongs to category 1.

\subsection{Analysis of Experimental Results}
In the experiments, the dataset was divided into training and test sets in an 8:2 ratio. Under the same experimental environment, keeping the parameters of the CIN component consistent with those in the xDeepFM model, comparative experiments were conducted on the learning rate and key hyperparameters in the multi-head attention mechanism. The performance of the proposed model was compared through four groups of experiments, discussing the impact of different hyperparameter combinations on the model's prediction results to obtain the optimal hyperparameter combination. During the experiments, AUC and Logloss were used to evaluate the model's results.

\subsubsection{Learning Rate Experiment}
In this experiment, we explored the impact of different learning rates on the model's prediction results. The model test results are shown in Figure 5. We selected eight different learning rates: 0.5, 0.1, 0.08, 0.06, 0.05, 0.001, 0.005, and 0.0001. Under the same number of training iterations, we compared the model's performance to evaluate their effectiveness during training. According to the AUC and Logloss results in Figure 5, the model achieved the best prediction performance when the learning rate was 0.05.

\begin{figure}[h!]
    \centering
    \includegraphics[width=0.8\textwidth]{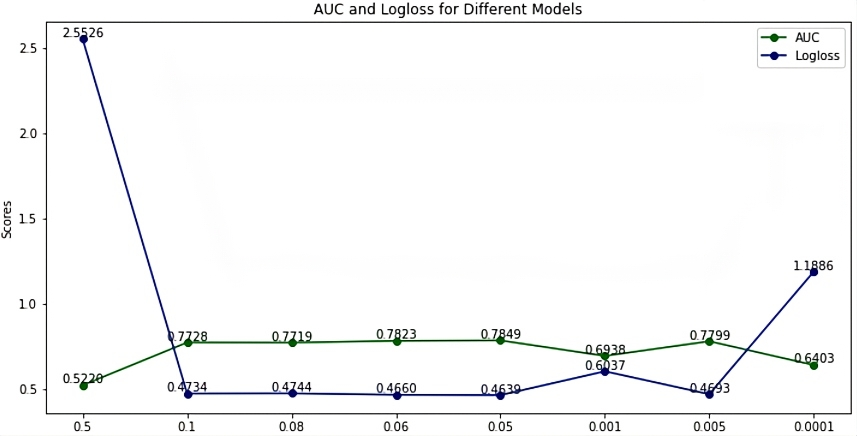}
    \caption{Summary of the impact of different learning rates on the predicted outcomes}
    \label{fig:example}
\end{figure}

\subsubsection{Embedding Vector Dimension Experiment}
In this experiment, we explored the impact of different embedding vector dimensions on the model's prediction results. The model test results are shown in Table 1 and Figure 6. Based on the experimental results in Section 3.3.1, we set the learning rate of the model to 0.05. Under the same number of training iterations and learning rate, we adjusted the embedding vector dimensions, conducting experiments with dimensions set to 2, 4, 6, and 8 sequentially, comparing the model's performance under different embedding dimensions to evaluate the model's effectiveness during training. According to the trends of AUC and Logloss in Figure 6, as the embedding vector dimension increased from 2 to 8, the AUC value increased, and the Logloss value decreased. When the embedding vector dimension was 8, the AUC value was the highest, and the Logloss value was the lowest. According to Table 1, when the embedding vector dimension was 8, the AUC value was 0.7849, and the Logloss value was 0.4629, indicating optimal prediction performance.

\begin{table}[h]
    \centering
    \begin{tabular}{ccc}
        \toprule
        \textbf{Embedding} & \textbf{AUC} & \textbf{Logloss} \\ \midrule
        2                  & 0.7819       & 0.4658           \\
        4                  & 0.7836       & 0.4641           \\
        6                  & 0.7844       & 0.4635           \\
        8                  & 0.7849       & 0.4629           \\ \bottomrule
    \end{tabular}
    \caption{Embedding size performance comparison}
    \label{tab:embedding_performance}
\end{table}

\begin{figure}[h!]
    \centering
    \includegraphics[width=0.8\textwidth]{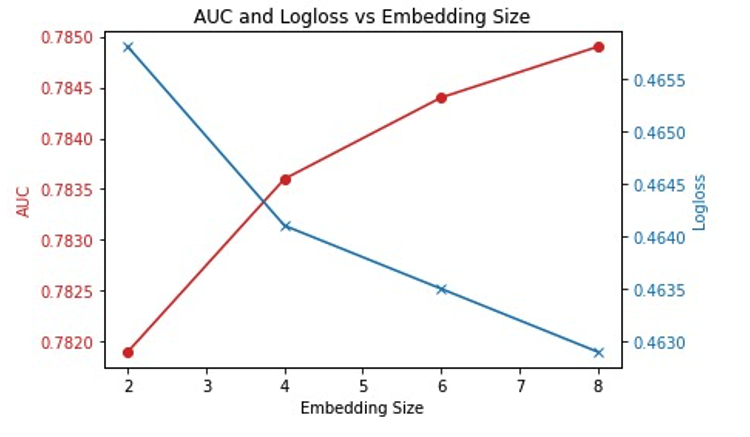}
    \caption{Effect of different Embedding vector dimensions on the prediction results}
    \label{fig:example}
\end{figure}

\subsubsection{Number of Attention Heads Experiment}
In this experiment, we explored the impact of different numbers of attention heads in the multi-head attention mechanism on the model's prediction results. The model test results are shown in Table 2 and Figure 7. Based on the experimental results in Sections 3.3.1 and 3.3.2, we set the learning rate to 0.05 and the embedding vector dimension to 8. Under the same number of training iterations, we adjusted the number of attention heads, conducting experiments with head counts set to 1, 2, 4, and 8 sequentially. According to the trends of AUC and Logloss in Figure 7, when the number of attention heads was 2, the AUC value was the highest, and the Logloss value was the lowest. Specifically, when the number of attention heads was 2, the AUC value was 0.7850, and the Logloss value was 0.4628, indicating optimal prediction performance.

\begin{figure}[h!]
    \centering
    \includegraphics[width=0.8\textwidth]{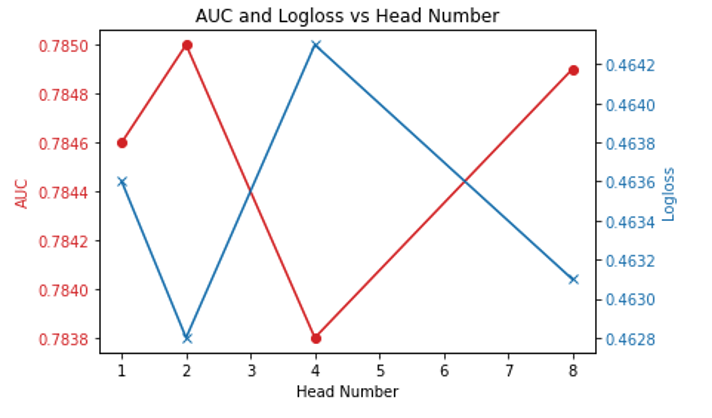}
    \caption{Effect of different attention head counts on predicted outcomes}
    \label{fig:example}
\end{figure}

\begin{table}[h]
    \centering
    \begin{tabular}{lcc}
        \toprule
        \textbf{Number of Long Attention Head} & \textbf{AUC} & \textbf{Logloss} \\ \midrule
        1                                      & 0.7846       & 0.4636           \\
        2                                      & 0.7850       & 0.4628           \\
        4                                      & 0.7838       & 0.4643           \\
        8                                      & 0.7849       & 0.4631           \\ \bottomrule
    \end{tabular}
    \caption{Performance comparison based on the number of long attention heads}
    \label{tab:attention_heads}
\end{table}

\subsubsection{Model Comparison Experiment}
To verify the effectiveness and accuracy of the proposed model on massive and sparse real advertising datasets, we compared our model with DeepFM and xDeepFM under the same experimental environment. In the experiments, the dataset was divided into training and test sets in an 8:2 ratio, and a total of 100 epochs were trained. The batch size was set to 2048 for the training set and 4096 for the test set. To ensure the accuracy of the experimental results, we used the same parameters to train each model: embedding dimension \cite{oh2016deep} was set to 8, the number of attention heads was set to 2, the hidden layer sizes of the DNN component were set to a two-layer decreasing structure of (256, 128), the optimizer was Adam, the learning rate was 0.05, and the learning scheduler was StepLR.
The test results of different models on the same dataset are shown in Table 3 and Figures 2 and 3. From the AUC and Logloss metrics, the proposed model outperformed the other two models. Compared with the xDeepFM model, our model's AUC value increased by 0.0072, and the Logloss value decreased by 0.0061. This comparison indicates that introducing the multi-head attention mechanism and replacing the linear model component with the FM model are effective strategies.
Additionally, we compared our model's test results with other existing models on the same dataset, and the results are as follows.

\begin{table}[h]
    \centering
    \begin{tabular}{lcc}
        \toprule
        \textbf{Model} & \multicolumn{2}{c}{\textbf{Criteo Datasets}} \\ \cmidrule(lr){2-3}
                       & \textbf{AUC} & \textbf{Logloss} \\ \midrule
        DeepFM         & 0.7816       & 0.4665           \\
        xDeepFM        & 0.7778       & 0.4689           \\
        SEDAFM         & 0.7757       & 0.4731           \\
        MAME-DFM       & 0.7791       & 0.4663           \\
        DTM            & 0.7625       & 0.4488           \\
        DMCNN          & 0.7994       & 0.5419           \\
        Paper Model    & 0.7850       & 0.4628           \\ \bottomrule
    \end{tabular}
    \caption{Performance comparison on Criteo Datasets}
    \label{tab:criteo_performance}
\end{table}

\section{Discussion}
In this paper, we present an advertisement click-through rate prediction model based on an improved xDeepFM model, aiming to improve the model performance and thus the accuracy of advertisement click-through rate, which in turn enhances the sustainable utilization of network resources. The effectiveness of the proposed improved model in advertisement click rate prediction is verified through experiments, and the results show that the model outperforms other similar models in two evaluation metrics, AUC and Logloss.
However, our work also has some limitations. First, the actual ad click rate prediction task is affected by a variety of factors, including user behavior, ad content, environmental factors, etc., and the complexity of these factors is beyond the consideration of our model. Second, the actual data may have problems such as noise, missing values or outliers, which may affect the training and prediction effect of the model. In addition, the dis-tribution of actual data changes with time and environment, so the model needs to be constantly updated and adapted to new data distributions and patterns.
In future research, we will more fully consider the challenges posed by the com-plexity of actual data to better reflect the complexity and diversity in real-world scenarios. By continuously optimizing the model structure and algorithms, we will improve the model's ability to fit and generalize complex data. Meanwhile, feature engineering and data preprocessing are also key, requiring in-depth research into methods to deal with issues such as noise, missing values and outliers in order to improve the quality and usability of data. In addition, we will explore techniques such as dynamic model updating and migration learning to adapt to changes in data distribution and patterns. By better understanding and solving the challenges in real data, we will further improve the performance and usefulness of ad click rate prediction models.

\section{Conclusion}
Due to the multi-level, nonlinear, and interactive relationships among features in advertising data, traditional methods often cannot effectively capture these complexities. The xDeepFM network, despite being a powerful feature interaction framework for CTR prediction, cannot effectively extract the correlations among user history, interests, and target advertisements. Its linear component mainly relies on shallow linear feature representations, failing to accurately capture differences and trends in user behavior across different time periods, which is not conducive to high-order feature learning. To improve the accuracy of CTR prediction, this paper proposes a CTR prediction model based on an improved xDeepFM model. We approach this from two perspectives: (1)By incorporating a multi-head attention mechanism into the original xDeepFM network model, the improved DNN component can better handle complex relationships among features. This mechanism computes feature relationships at different time points in a distributed manner and calculates internal attention parameters in different subspaces. It enhances the model's training effectiveness without increasing its computational complexity.(2)We replace the original linear model component of the xDeepFM model with the FM model, enabling the model to automatically capture first-order features and second-order features generated by pairwise combinations of first-order features. This allows the model to better handle high-dimensional discrete features and improves its ability to model highly sparse data.
We selected the Criteo dataset to explore the impact of different learning rates and numbers of attention heads on the experimental results and conducted comparative analyses between our proposed model and two other similar CTR prediction models. Experimental results show that the improved model achieved significant performance improvements compared to the original xDeepFM and DeepFM models on the same dataset, verifying the effectiveness of the proposed method. This enhancement not only increases the model's expressive capacity but also helps to better uncover the implicit relationships among features, further improving the model's prediction performance.

\bibliographystyle{unsrtnat}
\bibliography{references}  

\begin{thebibliography}{33}
\providecommand{\natexlab}[1]{#1}
\providecommand{\url}[1]{\texttt{#1}}
\expandafter\ifx\csname urlstyle\endcsname\relax
  \providecommand{\doi}[1]{doi: #1}\else
  \providecommand{\doi}{doi: \begingroup \urlstyle{rm}\Url}\fi

\bibitem[Sinclair(2020)]{sinclair2020cracking}
John Sinclair.
\newblock Cracking under pressure: current trends in the global advertising industry.
\newblock \emph{Media International Australia}, 174\penalty0 (1):\penalty0 3--16, 2020.

\bibitem[Sundaram et~al.(2020)Sundaram, Sharma, and Shakya]{sundaram2020power}
Rammohan Sundaram, Dr~Rajeev Sharma, and Dr~Anurag Shakya.
\newblock Power of digital marketing in building brands: A review of social media advertisement.
\newblock \emph{International Journal of Management}, 11\penalty0 (4), 2020.

\bibitem[Gudipudi et~al.(2023)Gudipudi, Nguyen, Bein, and Kurwadkar]{gudipudi2023improving}
Rakesh Gudipudi, Sandra Nguyen, Doina Bein, and Sudarshan Kurwadkar.
\newblock Improving internet advertising using click--through rate prediction.
\newblock \emph{Applied Human Factors and Ergonomics International}, 2023.

\bibitem[Gengler and Reynolds(1995)]{gengler1995consumer}
Charles~E Gengler and Thomas~J Reynolds.
\newblock Consumer understanding and advertising strategy: analysis and strategic translation of laddering data.
\newblock \emph{Journal of advertising research}, 35\penalty0 (4):\penalty0 19--34, 1995.

\bibitem[Evans(2009)]{evans2009online}
David~S Evans.
\newblock The online advertising industry: Economics, evolution, and privacy.
\newblock \emph{Journal of economic perspectives}, 23\penalty0 (3):\penalty0 37--60, 2009.

\bibitem[Wiles and Cornwell(1991)]{wiles1991review}
Judith~A Wiles and T~Bettina Cornwell.
\newblock A review of methods utilized in measuring affect, feelings, and emotion in advertising.
\newblock \emph{Current Issues and Research in Advertising}, 13\penalty0 (1-2):\penalty0 241--275, 1991.

\bibitem[Wuisan and Handra(2023)]{wuisan2023maximizing}
Dewi~Surya Wuisan and Tessa Handra.
\newblock Maximizing online marketing strategy with digital advertising.
\newblock \emph{Startupreneur Business Digital (SABDA Journal)}, 2\penalty0 (1):\penalty0 22--30, 2023.

\bibitem[Rong et~al.(2019)Rong, Xiao, Zhang, and Wang]{rong2019platform}
Ke~Rong, Fei Xiao, Xiaoyu Zhang, and Jingjing Wang.
\newblock Platform strategies and user stickiness in the online video industry.
\newblock \emph{Technological Forecasting and Social Change}, 143:\penalty0 249--259, 2019.

\bibitem[Hosmer~Jr et~al.(2013)Hosmer~Jr, Lemeshow, and Sturdivant]{hosmer2013applied}
David~W Hosmer~Jr, Stanley Lemeshow, and Rodney~X Sturdivant.
\newblock \emph{Applied logistic regression}.
\newblock John Wiley \& Sons, 2013.

\bibitem[Feng et~al.(2018)Feng, Yu, and Zhou]{feng2018multi}
Ji~Feng, Yang Yu, and Zhi-Hua Zhou.
\newblock Multi-layered gradient boosting decision trees.
\newblock \emph{Advances in neural information processing systems}, 31, 2018.

\bibitem[Rendle(2010)]{rendle2010factorization}
Steffen Rendle.
\newblock Factorization machines.
\newblock In \emph{2010 IEEE International conference on data mining}, pages 995--1000. IEEE, 2010.

\bibitem[He et~al.(2014)He, Pan, Jin, Xu, Liu, Xu, Shi, Atallah, Herbrich, Bowers, et~al.]{he2014practical}
Xinran He, Junfeng Pan, Ou~Jin, Tianbing Xu, Bo~Liu, Tao Xu, Yanxin Shi, Antoine Atallah, Ralf Herbrich, Stuart Bowers, et~al.
\newblock Practical lessons from predicting clicks on ads at facebook.
\newblock In \emph{Proceedings of the eighth international workshop on data mining for online advertising}, pages 1--9, 2014.

\bibitem[Gai et~al.(2017)Gai, Zhu, Li, Liu, and Wang]{gai2017learning}
Kun Gai, Xiaoqiang Zhu, Han Li, Kai Liu, and Zhe Wang.
\newblock Learning piece-wise linear models from large scale data for ad click prediction.
\newblock \emph{arXiv preprint arXiv:1704.05194}, 2017.

\bibitem[Wiriyathammabhum et~al.(2016)Wiriyathammabhum, Summers-Stay, Ferm{\"u}ller, and Aloimonos]{wiriyathammabhum2016computer}
Peratham Wiriyathammabhum, Douglas Summers-Stay, Cornelia Ferm{\"u}ller, and Yiannis Aloimonos.
\newblock Computer vision and natural language processing: recent approaches in multimedia and robotics.
\newblock \emph{ACM Computing Surveys (CSUR)}, 49\penalty0 (4):\penalty0 1--44, 2016.

\bibitem[Mahesh(2020)]{mahesh2020machine}
Batta Mahesh.
\newblock Machine learning algorithms-a review.
\newblock \emph{International Journal of Science and Research (IJSR).[Internet]}, 9\penalty0 (1):\penalty0 381--386, 2020.

\bibitem[Hochreiter(1997)]{hochreiter1997long}
S~Hochreiter.
\newblock Long short-term memory.
\newblock \emph{Neural Computation MIT-Press}, 1997.

\bibitem[O'Shea(2015)]{o2015introduction}
K~O'Shea.
\newblock An introduction to convolutional neural networks.
\newblock \emph{arXiv preprint arXiv:1511.08458}, 2015.

\bibitem[Bebis and Georgiopoulos(1994)]{bebis1994feed}
George Bebis and Michael Georgiopoulos.
\newblock Feed-forward neural networks.
\newblock \emph{Ieee Potentials}, 13\penalty0 (4):\penalty0 27--31, 1994.

\bibitem[Cheng et~al.(2016)Cheng, Koc, Harmsen, Shaked, Chandra, Aradhye, Anderson, Corrado, Chai, Ispir, et~al.]{cheng2016wide}
Heng-Tze Cheng, Levent Koc, Jeremiah Harmsen, Tal Shaked, Tushar Chandra, Hrishi Aradhye, Glen Anderson, Greg Corrado, Wei Chai, Mustafa Ispir, et~al.
\newblock Wide \& deep learning for recommender systems.
\newblock In \emph{Proceedings of the 1st workshop on deep learning for recommender systems}, pages 7--10, 2016.

\bibitem[Guo et~al.(2017)Guo, Tang, Ye, Li, and He]{guo2017deepfm}
Huifeng Guo, Ruiming Tang, Yunming Ye, Zhenguo Li, and Xiuqiang He.
\newblock Deepfm: a factorization-machine based neural network for ctr prediction.
\newblock \emph{arXiv preprint arXiv:1703.04247}, 2017.

\bibitem[Lian et~al.(2018)Lian, Zhou, Zhang, Chen, Xie, and Sun]{lian2018xdeepfm}
Jianxun Lian, Xiaohuan Zhou, Fuzheng Zhang, Zhongxia Chen, Xing Xie, and Guangzhong Sun.
\newblock xdeepfm: Combining explicit and implicit feature interactions for recommender systems.
\newblock In \emph{Proceedings of the 24th ACM SIGKDD international conference on knowledge discovery \& data mining}, pages 1754--1763, 2018.

\bibitem[Wang et~al.(2017)Wang, Fu, Fu, and Wang]{wang2017deep}
Ruoxi Wang, Bin Fu, Gang Fu, and Mingliang Wang.
\newblock Deep \& cross network for ad click predictions.
\newblock In \emph{{Proceedings of the ADKDD'17}}, pages 1--7. 2017.

\bibitem[Wang et~al.(2019)Wang, Jiang, Hou, Huang, Dou, Zhang, and Guo]{wang2019survey}
Zhengjie Wang, Kangkang Jiang, Yushan Hou, Zehua Huang, Wenwen Dou, Chengming Zhang, and Yinjing Guo.
\newblock A survey on csi-based human behavior recognition in through-the-wall scenario.
\newblock \emph{IEEE Access}, 7:\penalty0 78772--78793, 2019.

\bibitem[Okada et~al.(2019)Okada, Ohzeki, and Taguchi]{okada2019efficient}
Shuntaro Okada, Masayuki Ohzeki, and Shinichiro Taguchi.
\newblock Efficient partition of integer optimization problems with one-hot encoding.
\newblock \emph{Scientific reports}, 9\penalty0 (1):\penalty0 13036, 2019.

\bibitem[Zhang et~al.(2021)Zhang, Qin, Guo, Tang, and He]{zhang2021deep}
Weinan Zhang, Jiarui Qin, Wei Guo, Ruiming Tang, and Xiuqiang He.
\newblock Deep learning for click-through rate estimation.
\newblock \emph{arXiv preprint arXiv:2104.10584}, 2021.

\bibitem[Song et~al.(2019)Song, Shi, Xiao, Duan, Xu, Zhang, and Tang]{song2019autoint}
Weiping Song, Chence Shi, Zhiping Xiao, Zhijian Duan, Yewen Xu, Ming Zhang, and Jian Tang.
\newblock Autoint: Automatic feature interaction learning via self-attentive neural networks.
\newblock In \emph{Proceedings of the 28th ACM international conference on information and knowledge management}, pages 1161--1170, 2019.

\bibitem[Vaswani(2017)]{vaswani2017attention}
A~Vaswani.
\newblock Attention is all you need.
\newblock \emph{Advances in Neural Information Processing Systems}, 2017.

\bibitem[Wang et~al.(2022)Wang, Fang, Zhong, Zhuo, Peng, and Xu]{wang2022spatial}
Yuxian Wang, Yuan Fang, Wenlong Zhong, Rongming Zhuo, Junhuan Peng, and Linlin Xu.
\newblock A spatial--temporal depth-wise residual network for crop sub-pixel mapping from modis images.
\newblock \emph{Remote Sensing}, 14\penalty0 (21):\penalty0 5605, 2022.

\bibitem[Deng et~al.(2021)Deng, Pan, Zhou, Kong, Flores, and Lin]{deng2021deeplight}
Wei Deng, Junwei Pan, Tian Zhou, Deguang Kong, Aaron Flores, and Guang Lin.
\newblock Deeplight: Deep lightweight feature interactions for accelerating ctr predictions in ad serving.
\newblock In \emph{Proceedings of the 14th ACM international conference on Web search and data mining}, pages 922--930, 2021.

\bibitem[Lobo et~al.(2008)Lobo, Jim{\'e}nez-Valverde, and Real]{lobo2008auc}
Jorge~M Lobo, Alberto Jim{\'e}nez-Valverde, and Raimundo Real.
\newblock Auc: a misleading measure of the performance of predictive distribution models.
\newblock \emph{Global ecology and Biogeography}, 17\penalty0 (2):\penalty0 145--151, 2008.

\bibitem[Bradley(1997)]{bradley1997use}
Andrew~P Bradley.
\newblock The use of the area under the roc curve in the evaluation of machine learning algorithms.
\newblock \emph{Pattern recognition}, 30\penalty0 (7):\penalty0 1145--1159, 1997.

\bibitem[Bishop(1995)]{bishop1995neural}
Christopher~M Bishop.
\newblock \emph{Neural networks for pattern recognition}.
\newblock Oxford university press, 1995.

\bibitem[Oh~Song et~al.(2016)Oh~Song, Xiang, Jegelka, and Savarese]{oh2016deep}
Hyun Oh~Song, Yu~Xiang, Stefanie Jegelka, and Silvio Savarese.
\newblock Deep metric learning via lifted structured feature embedding.
\newblock In \emph{Proceedings of the IEEE conference on computer vision and pattern recognition}, pages 4004--4012, 2016.

\end{thebibliography}

\end{document}